\documentclass{article}

\usepackage{natbib}
\usepackage{hyperref}
\usepackage[labelfont=bf]{caption}
\hypersetup{
  colorlinks,
  citecolor=black,
  filecolor=black,
  linkcolor=black,
  urlcolor=black
}
\usepackage[margin=1in]{geometry}
\usepackage{amsmath, amsfonts, bm}
\usepackage{graphicx}
\usepackage{epstopdf}
\usepackage{cleveref} 
\usepackage{setspace} 
\usepackage{xfrac}
\usepackage{caption}
\usepackage{subcaption}
\usepackage{booktabs}
\usepackage{placeins}
\usepackage{algorithmic}
\ifpdf
  \DeclareGraphicsExtensions{.eps,.pdf,.png,.jpg}
\else
  \DeclareGraphicsExtensions{.eps}
\fi

\usepackage{xcolor}

\title{Efficient Closure Learning in Dynamical Systems: A Comparison of Data-Assimilated and Embedded Approaches}

\usepackage[affil-it]{authblk} 

\author[1]{Rileigh Bandy}
\author[1]{Teresa Portone\thanks{Corresponding author: tporton@sandia.gov}}
\author[1]{Mohammad Khalil}
\author[2]{Eric Crislip}
\author[1]{Kyle Neal}

\affil[1]{Sandia National Laboratories}
\affil[2]{Department of Statistics, The Ohio State University, Columbus, Ohio}

\usepackage{amsopn}

\usepackage{subcaption}
\usepackage{makecell}
\usepackage{todonotes}
\usepackage{amsmath}
\DeclareMathOperator*{\argmin}{argmin}

\begin{document}

\maketitle

\begin{abstract}
Increased computing power has transformed computational science and improved predictive capabilities for complex systems, but many phenomena still exceed current simulation capabilities or require rapid evaluations for fast design iterations.
Closure models address this challenge by approximating unresolved effects in reduced or incomplete dynamical systems.
Data-driven closures offer a flexible mechanism for improving model accuracy and applicability, but approaches that embed the closure within a computational model for construction and training phases can be computationally expensive and cause numerical instabilities in the computational model during training.
To avoid the challenges posed by embedded approaches, staged approaches to closure-model learning have been developed.
The first stage uses data assimilation to infer state and closure trajectories from observations.
The second stage trains a machine learning closure model offline using the assimilated data, thereby avoiding repeated simulator calls during training.
In this work, we investigate the cost and accuracy tradeoffs between embedded and staged data-driven closure modeling approaches for a synthetic epidemiological SIQR model with a nonlinear state-dependent quarantine closure. 
The results illustrate a practical accuracy-efficiency trade-off: for this exemplar, embedded training provides the strongest average accuracy, while the staged approach yields comparable qualitative behavior with a moderate increase in error, substantially fewer model evaluations, and improved optimization robustness.
\end{abstract}



\section{Introduction}

Increased computational power has enabled high-fidelity simulation across a wide range of scientific and engineering applications, but significant challenges remain in constructing accurate and complete mathematical models.
A central issue is the \emph{closure problem}, in which unknown quantities appear in the governing equations and must be modeled in order to close the system.
These unknown quantities may arise from unresolved scales, incomplete physics, or coarse-grained representations.
Prominent examples of closure problems include Reynolds stresses in Reynolds-averaged Navier--Stokes equations, subgrid-scale stresses in large eddy simulations, and constitutive relations in continuum mechanics \cite{srivastava2021generalizable,durbin2018some,hanjalic1994advanced}.

Traditional approaches to constructing closure models rely on dimensional analysis, asymptotic reasoning, scaling arguments, and empirical calibration \cite{bergel2023neural, durbin2018some, hanjalic1994advanced, ray2021verification}.
These methods have produced robust and widely used models that perform well within specific calibration regimes.
However, they often result in relatively simple algebraic or low-dimensional functional forms, which can limit expressiveness and adaptability.
Consequently, such closures may struggle to generalize across regimes or capture complex nonlinear interactions in many physical and biological systems \cite{pan2018data}.

These limitations motivate more flexible data-driven approaches to closure modeling.
Recent advances in machine learning, together with increasing availability of high-fidelity simulation and experimental data, have enabled the construction of closure models based on expressive function approximators such as neural networks~\cite{parish_paradigm_2016,pan_data-driven_2018,gupta_neural_2021,porteous_data-driven_2023,huang_machine_2022,rasp_deep_2018,parish2023turbulence,sanderse_scientific_2025}.
Such approaches can capture nonlinear dependencies and latent structures that are difficult to encode using traditional closure modeling paradigms.
However, significant challenges remain when neural closures are trained while embedded directly within the governing equations.
Each loss or validation evaluation requires a forward model solve, and training may require differentiating through the numerical solver.
Embedded training can also be numerically fragile: as the neural-network parameters evolve, the learned closure may alter the system dynamics in ways that make the initial-value problem difficult to solve, for example by inducing stiffness, nonphysical states, or extremely small stable time steps.
These challenges become more pronounced when multiple training trajectories or parameter scenarios are used to improve generalizability, because each scenario requires an additional model solve during optimization. 

To address these challenges, staged approaches to data-driven closure modeling have been developed~\cite{neal2024investigating, crislip2025closuretermestimationspatiotemporal, subramanian2019bayesian, subramanian2019error, subramanian2020model}.
The central idea is to decouple closure inference from closure-function training.
In the first stage, data assimilation is used to infer closure-consistent state and closure trajectories from observations by augmenting the physical state with a latent closure variable. 
In the second stage, a regression or machine-learning model is trained offline to map resolved states to the inferred closure values. 
After training, the learned closure can be embedded into the governing equations for prediction. 
This workflow avoids repeated forward simulations during neural-network optimization and does not require the simulator to be differentiable.

Despite the practical appeal of staged closure learning, the accuracy--cost trade-offs relative to embedded neural-closure training are not yet fully understood. 
This paper therefore presents a systematic comparison of these tradeoffs in the context of a synthetic compartment-based epidemiological model, where an unknown closure represents a state-dependent transition rate from infectious to quarantined populations.
Because the study is synthetic, the true closure is known and can be used for diagnostic evaluation, although the methods themselves use only noisy observations of population state variables. 
For this study, we find that the embedded approach achieves lower average held-out calibration error and lower average calibration closure error, but requires $9\times$ greater model evaluations for training than the staged approach.

The remainder of the paper is organized as follows.
Section~\ref{sec:background} reviews the closure-learning problem, embedded neural closure modeling, and the data-assimilation tools used for staged closure inference.
Section~\ref{sec:methodology} defines the comparison protocol between embedded and staged approaches, including data usage, neural-network optimization, performance metrics, computational-cost accounting, and robustness measures.
Section~\ref{sec:numerical} presents the synthetic SIQR numerical study and compares the two methods in terms of held-out calibration accuracy, diagnostic closure error, training-stage model evaluations, failed neural-network starts, and validation-scenario generalization.
Section~\ref{sec:conclusions} summarizes the main findings and discusses future directions.
\section{Background}
\label{sec:background}

\subsection{The closure problem}
\label{sec:background_closure_problem}

Many systems of interest can be represented as dynamical systems whose resolved state $x(t) \in \mathbb{R}^{n_x}$ evolves according to
\begin{equation}
    \frac{d x(t)}{d t} = R(x(t), \xi(x(t))),
    \label{eq:true_closed_dynamics}
\end{equation}
where $R$ denotes the known resolved dynamics and $\xi$ denotes an unknown closure.
The closure may represent unresolved physics, model-form error, coarse-graining effects, missing source terms, or unresolved couplings.
The closure problem arises because the governing equations cannot be evaluated without specifying $\xi$.
In many applications, the functional form of $\xi$ is unavailable or only approximately known.

The goal of data-driven closure learning is to construct a parameterized approximation $\xi_\theta$ from observations of the resolved state or other derived quantities of interest.
Once learned, the closure model is embedded into the dynamical system to produce the closed predictive model
\begin{equation}
    \frac{d x(t)}{d t} = R(x(t), \xi_\theta(x(t))).
    \label{eq:learned_closed_dynamics}
\end{equation}

Suppose observations are available at times $\{t_i\}_{i=0}^{N_t-1}$.
A linear observation model can be defined as
\begin{equation}
    d_i = H x(t_i) + \epsilon_i,
    \qquad
    \epsilon_i \overset{iid}{\sim} \mathcal{N}(0,\Gamma),
    \label{eq:background_observation_model_physical}
\end{equation}
where $H$ maps the resolved state to the observed quantities and $\Gamma$ is the observation-error covariance.
For any given training procedure, a subset of all observation times may be employed (e.g., for test-train-validation split). 
In the following discussion we denote a subset of observation-time indices as $\mathcal{I}$, and we specify which indices are used for different training procedures in detail in~\Cref{sec:methodology}.


\subsection{Embedded training}
\label{sec:background_embedded_training}

Embedded training places the parametric closure model directly inside the governing equations during optimization.
For a neural-network closure $\xi_\theta$, the embedded model is
\begin{equation}
    \frac{d x(t;\theta)}{d t}
    =
    R(x(t;\theta), \xi_\theta(x(t;\theta))),
    \label{eq:embedded_training_dynamics}
\end{equation}
where $x(t;\theta)$ denotes the model trajectory obtained with closure parameters $\theta$.
For Gaussian observation errors, embedded training minimizes the negative log-likelihood of the training observations.
Up to additive constants independent of $\theta$, this objective is
\begin{equation}
    \mathcal{L}_{\mathrm{emb}}(\theta)
    =
    \frac{1}{2}
    \sum_{i \in \mathcal{I}}
    \left(
        d_i - H x(t_i;\theta)
    \right)^\top
    \Gamma^{-1}
    \left(
        d_i - H x(t_i;\theta)
    \right).
    \label{eq:embedded_training_loss_background}
\end{equation}

This approach is closely related to universal differential equations, which augment mechanistic models with trainable function approximators~\cite{rackauckas2020universal}.
A key advantage of embedded training is that the closure is optimized according to its direct effect on the observable state trajectory.
However, each loss or validation evaluation requires a forward solve of the dynamical system.
Gradient-based optimization may also require differentiating through the numerical time integrator.
These costs become more pronounced when the data includes multiple scenarios (e.g., multiple initial conditions), because each scenario requires a unique model solve.
Embedded training can also be numerically fragile because intermediate neural-network parameters may alter the dynamics in ways that make the initial-value problem difficult to solve.
For example, the learned closure may induce stiffness, nonphysical states, exploding trajectories, or extremely small time steps to maintain numerical stability.

\subsection{Staged training}
\label{sec:background_staged_training}

Staged training separates closure inference from closure-function training.
The first stage uses data assimilation to infer a time-dependent closure trajectory that is consistent with the observations.
The second stage trains a regression model offline to map resolved states to the inferred closure values.
After this supervised training step, the learned closure is embedded into the governing equations for prediction.
Thus, staged training invokes the expensive computational model during data assimilation, but avoids repeated model calls during neural-network optimization.
It also does not require the computational model to be differentiable with respect to closure-model parameters.

\subsubsection{Step 1: Data assimilation by state augmentation, EnKF, and smoothing}
\label{sec:background_staged_step1}

The staged approach considered here estimates the latent closure by augmenting the physical state with the closure variable.
Define the augmented state as
\begin{equation}
    z(t)
    =
    \begin{bmatrix}
        x(t) \\
        \xi(t)
    \end{bmatrix}.
    \label{eq:augmented_state_background}
\end{equation}
The physical state evolves according to the known model evaluated with a time-dependent closure, while the closure is assigned artificial stochastic dynamics:
\begin{align}
    \frac{d x(t)}{d t} &= R(x(t), \xi(t)),
    \label{eq:augmented_physics_background} \\
    d \xi(t) &= \sigma_\xi \, dW(t).
    \label{eq:closure_brownian_background}
\end{align}
Here $W(t)$ is a Wiener process and $\sigma_\xi$ controls the closure process-noise magnitude.
The stochastic closure dynamics are not intended to be a mechanistic model for the closure.
Instead, they provide flexibility for the data-assimilation procedure to attribute model-data mismatch to an unresolved closure process.

The observation model for the augmented state is
\begin{equation}
    d_i = \mathcal{H} z(t_i) + \epsilon_i,
    \qquad
    \epsilon_i \overset{iid}{\sim} \mathcal{N}(0,\Gamma).
    \label{eq:background_observation_model_augmented}
\end{equation}
If only the physical state is observed, then $\mathcal{H} = \begin{bmatrix} H & 0 \end{bmatrix}$.

We use an ensemble Kalman filter (EnKF)~\cite{evensen2003ensemble} to sequentially update an ensemble representation of the augmented state.
At each observation time, the ensemble is first propagated through the augmented dynamics in~\eqref{eq:augmented_physics_background}--\eqref{eq:closure_brownian_background}.
The forecast ensemble is then corrected using the discrepancy between the predicted and observed data.
Because the closure variable is part of the augmented state, this update adjusts both the resolved physical variables and the latent closure through their ensemble-estimated correlations with the observations.
The closure process-noise scale $\sigma_\xi$ can be fixed or estimated from the observations used in the assimilation window.
In this work, $\sigma_\xi$ is selected by maximizing an EnKF innovation likelihood, following the innovation-based covariance-parameter estimation approach of \cite{dee1995line}.

Filtering estimates at time \(t_i\) use observations only up to that time.
For closure learning, however, it is useful for the inferred closure labels to reflect information from the full assimilated time window.
We therefore apply a fixed-interval ensemble smoother~\cite{evensen_ensemble_2000}.
The smoother retrospectively updates stored ensembles at previous output times using later observation innovations and ensemble-estimated temporal correlations.
After all observations in the assimilation window have been processed, the resulting smoothed ensemble provides closure-consistent estimates of both the resolved state and the latent closure.

\subsubsection{Step 2: Offline closure training}
\label{sec:background_staged_step2}

The full smoothed ensemble contains more information than its mean.
In particular, the ensemble spread provides an estimate of uncertainty in both the resolved state and the inferred closure.
This uncertainty could be propagated into the learned closure model by training an uncertainty-aware regressor on the ensemble rather than only on the ensemble mean.
For example, the ensemble at each time could be used to define a local likelihood for Bayesian neural-network training, or multiple closure networks could be trained on ensemble-derived samples to form an ensemble closure model.
Such approaches would allow uncertainty in the data-assimilated closure estimates to be carried into the final autonomous predictions.
In this work, however, we use only the smoothed ensemble mean.
This deterministic reduction is chosen to maintain a direct comparison with the deterministic embedded neural-network closure.

The data-assimilated closure-training pairs are
\begin{equation}
    \mathcal{D}_{\mathrm{DA}}
    =
    \left\{
        \left(\bar{x}_i^s,\bar{\xi}_i^s\right)
        :
        i \in \mathcal{I}
    \right\},
    \label{eq:da_closure_training_set_background}
\end{equation}
where \(\bar{x}_i^s\) and \(\bar{\xi}_i^s\) denote the smoothed ensemble means of the resolved state and inferred closure, respectively.
A neural network \(\xi_\theta\) is then trained offline to approximate the map from resolved state to inferred closure:
\begin{equation}
    \theta^* = \argmin_\theta
    \frac{1}{2}
    \sum_{i \in \mathcal{I}}
    \left|
        \xi_\theta(\bar{x}_i^s) - \bar{\xi}_i^s
    \right|^2.
    \label{eq:staged_supervised_objective_background}
\end{equation}
After training, the learned closure is inserted into the original dynamical model for new predictions:
\begin{equation}
    \frac{d x(t)}{d t}
    =
    R(x(t), \xi_{\theta^*}(x(t))).
    \label{eq:staged_final_rollout_background}
\end{equation}
\section{Comparison Methodology}
\label{sec:methodology}

This section defines the experimental protocol used to compare embedded closure training with staged data-assimilated closure training.
The general mathematical approach for both methods is given in Section~\ref{sec:background}.
Here, we specify how the observation data are split, how each method uses those data, how neural-network training is performed, and how accuracy, cost, and robustness are measured.

\subsection{Data usage and held-out evaluation}
\label{sec:methodology_data_usage}

Both methods are evaluated using the same noisy observations and the same randomized train, validation, and test time-index split.
The observation-time indices are partitioned into $\mathcal{I}_{\mathrm{train}}$, $\mathcal{I}_{\mathrm{val}}$, and $\mathcal{I}_{\mathrm{test}}$.
The split fractions are $60\%$, $20\%$, and $20\%$, respectively.
The test indices are held out from all training, data assimilation, hyperparameter selection, early stopping, and model-selection procedures.

For embedded training, the neural closure is trained directly against the noisy observations indexed by $\mathcal{I}_{\mathrm{train}}$.
The validation observations indexed by $\mathcal{I}_{\mathrm{val}}$ are used for early stopping, random-start selection, and derivative-based local refinement accept-or-reject decisions.
The test observations indexed by $\mathcal{I}_{\mathrm{test}}$ are used only for final evaluation.

For staged training, data assimilation is performed once using the union of the original training and validation observations.
Thus, the assimilation index set is
\begin{equation}
    \mathcal{I}_{\mathrm{assim}}
    =
    \mathcal{I}_{\mathrm{train}}
    \cup
    \mathcal{I}_{\mathrm{val}}.
    \label{eq:assimilation_index_set_methods}
\end{equation}
The original test observations are never assimilated.
The closure process-noise parameter $\sigma_\xi$ is estimated using only observations indexed by $\mathcal{I}_{\mathrm{assim}}$.
After the EnKF and smoother are run, pseudo-training pairs are selected at the original training times and pseudo-validation pairs are selected at the original validation times:
\begin{align}
    \mathcal{D}_{\mathrm{pseudo\text{-}train}}
    &=
    \left\{
        \left(\bar{x}_i^s,\bar{\xi}_i^s\right)
        :
        i \in \mathcal{I}_{\mathrm{train}}
    \right\},
    \label{eq:pseudo_train_methods}
    \\
    \mathcal{D}_{\mathrm{pseudo\text{-}val}}
    &=
    \left\{
        \left(\bar{x}_i^s,\bar{\xi}_i^s\right)
        :
        i \in \mathcal{I}_{\mathrm{val}}
    \right\}.
    \label{eq:pseudo_val_methods}
\end{align}
The prefix ``pseudo'' emphasizes that the supervised learning targets \(\bar{\xi}_i^s\) are not directly observed closure values, but are inferred quantities produced by the EnKF and smoother.
The staged approach uses the original train--validation index split to organize these inferred pairs, but the pseudo-training and pseudo-validation sets are not independent raw datasets.
Because both sets are produced by a single assimilation and smoothing procedure over \(\mathcal{I}_{\mathrm{assim}}\), they are temporally correlated through the forecast dynamics, ensemble covariances, and retrospective smoothing updates.
Thus, pseudo-validation is used only for internal staged neural-network selection, such as early stopping or random-start selection decisions.
It is not treated as an independent measure of predictive performance.
Final accuracy comparisons are made only on the original held-out observations indexed by \(\mathcal{I}_{\mathrm{test}}\), which are excluded from assimilation, training, validation, and model selection.
\subsection{Closure-model training objectives}
\label{sec:methodology_training_objectives}

Both approaches use the same parametric closure-model class.
They differ in the objective used to construct the closure and in whether the dynamical-system simulator is invoked during closure-model training.

In the embedded approach, the closure is trained according to its effect on the observed state trajectory.
For a candidate closure \(\xi_\theta\), the model trajectory \(x(t;\theta)\) is obtained by solving~\eqref{eq:embedded_training_dynamics}.
The embedded training objective is the observation-space negative log-likelihood over the training indices, up to additive constants independent of \(\theta\):
\begin{equation}
    \mathcal{L}_{\mathrm{emb}}(\theta)
    =
    \frac{1}{2}
    \sum_{i \in \mathcal{I}_{\mathrm{train}}}
    \left(
        d_i - H x(t_i;\theta)
    \right)^T
    \Gamma^{-1}
    \left(
        d_i - H x(t_i;\theta)
    \right).
    \label{eq:methodology_embedded_loss}
\end{equation}
Multistart and stopping criteria evaluations use the analogous observation-space loss over  \(\mathcal{I}_{\mathrm{val}}\).

In the staged approach, the closure is trained as an offline supervised regression problem using the data-assimilated state--closure pairs.
The staged training objective is
\begin{equation}
    \mathcal{L}_{\mathrm{staged}}(\theta)
    =
    \frac{1}{2}
    \sum_{i \in \mathcal{I}_{\mathrm{train}}}
    \left|
        \xi_\theta(\bar{x}_i^s) - \bar{\xi}_i^s
    \right|^2.
    \label{eq:methodology_staged_loss}
\end{equation}
Multistart and stopping criteria evaluations use the analogous supervised loss over \(\mathcal{D}_{\mathrm{pseudo\text{-}val}}\).
This supervised training step does not call the dynamical-system simulator.
After training, the selected staged closure is embedded into the governing equations for future predictions.

\subsection{Performance metrics}
\label{sec:methodology_metrics}

Both approaches are evaluated on the same held-out test observations.
The primary calibration metric is the likelihood-scaled observation space test loss
\begin{equation}
    \mathcal{L}_{\mathrm{test}}
    =
    \frac{1}{2}
    \sum_{i \in \mathcal{I}_{\mathrm{test}}}
    \left(
        d_i - H x_{\mathrm{pred}}(t_i)
    \right)^T
    \Gamma^{-1}
    \left(
        d_i - H x_{\mathrm{pred}}(t_i)
    \right),
    \label{eq:methodology_test_loss}
\end{equation}
where \(x_{\mathrm{pred}}(t_i)\) denotes the state trajectory produced by the trained closure model.
This metric evaluates predictive agreement with raw observations that were not used during training, data assimilation, validation, or model selection.

Because the numerical study is synthetic, the true closure is known and can be used for diagnostic evaluation.
For an index set \(\mathcal{I}\), the closure error is measured by
\begin{equation}
    \mathrm{RMSE}_{\xi}(\mathcal{I})
    =
    \left(
        \frac{1}{|\mathcal{I}|}
        \sum_{i \in \mathcal{I}}
        \left|
            \xi_\theta(x_{\mathrm{true}}(t_i))
            -
            \xi_{\mathrm{true}}(x_{\mathrm{true}}(t_i))
        \right|^2
    \right)^{1/2}.
    \label{eq:methodology_closure_rmse}
\end{equation}
This closure RMSE is a diagnostic metric and would not be available in a real closure-learning application where \(\xi_{\mathrm{true}}\) is unknown.

To assess generalization, trained closures are also embedded into systems with parameter values different from those used during calibration.
For each of these validation scenarios, state prediction error is measured by
\begin{equation}
    \mathrm{RMSE}_{x}
    =
    \left(
        \frac{1}{N_t}
        \sum_{i=0}^{N_t-1}
        \left\|
            x_{\mathrm{pred}}(t_i)
            -
            x_{\mathrm{true}}(t_i)
        \right\|_2^2
    \right)^{1/2}.
    \label{eq:methodology_state_rmse}
\end{equation}
Closure prediction error in each validation scenario is measured analogously using \(\mathrm{RMSE}_{\xi}\) over the validation trajectory.

\subsection{Computational cost and robustness accounting}
\label{sec:methodology_cost_robustness}

Computational cost is measured by the number of computational model evaluations required to construct each closure model.
This accounting is intended to isolate the portion of each workflow that would become expensive for large, legacy, or high-fidelity simulators.

For embedded training, the cost includes all forward solves of the dynamical system used during closure-model optimization and model selection.
Each evaluation of the embedded training or validation objective requires a computational model evaluation with the current closure parameters.
Thus, embedded cost scales directly with the number of candidate closure models evaluated during training.

For staged training, the cost includes computational model evaluations used during the data-assimilation stage, including any evaluations used to select data-assimilation hyperparameters and the final assimilation and smoothing pass.
The subsequent supervised closure-regression step is not included in the total cost because it evaluates only the offline regression objective on the assimilated state--closure pairs.
This distinction reflects the central computational difference between the two workflows: embedded training repeatedly calls the simulator during neural-network optimization, whereas staged training shifts simulator use into a preprocessing stage and then performs closure-function training offline.

Robustness is measured by the number of failed candidate training runs.
A candidate is counted as failed if it produces a solver failure, nonfinite state, or nonfinite loss.
This metric is included because embedded optimization can generate intermediate closures that alter the dynamical system in unstable or nonphysical ways.
In contrast, after the data-assimilation stage, the staged neural-network optimization is a supervised regression problem, so unstable intermediate closure models do not impact computational model performance during training.

\subsection{Replicate analysis}
\label{sec:replicate_methodology}
To assess the variability in comparisons of accuracy and cost, we perform replicate trainings herein, where each replicate features a new noisy observation realization, a new random train-validation-test split, and a new seed for multistart neural-network training. 
Both methods are evaluated on the same held-out test observations per replicate realization.
These replicate results are aggregated and presented in the numerical results in~\Cref{sec:numerical}.
\section{Numerical study}
\label{sec:numerical}

\subsection{SIQR epidemiological exemplar}
\label{sec:siqr_exemplar}

We compare embedded and staged closure training using a synthetic compartmental epidemiological model with susceptible, infectious, quarantined, and removed compartments.
The state is
\begin{equation}
    x(t)
    =
    \begin{bmatrix}
        s(t) & i(t) & q(t) & r(t)
    \end{bmatrix}^T,
    \label{eq:siqr_state}
\end{equation}
where each component is a population fraction.
The true dynamics are
\begin{align}
    \frac{d s}{d t}
    &=
    -\beta s i,
    \label{eq:true_siqr_s}
    \\
    \frac{d i}{d t}
    &=
    \beta s i - \gamma i - \xi(s,i)i,
    \label{eq:true_siqr_i}
    \\
    \frac{d q}{d t}
    &=
    \xi(s,i)i - \delta q,
    \label{eq:true_siqr_q}
    \\
    \frac{d r}{d t}
    &=
    \gamma i + \delta q.
    \label{eq:true_siqr_r}
\end{align}
The unknown closure $\xi$ controls the transition from infectious to quarantined individuals.
The true synthetic closure is
\begin{equation}
    \xi(s(t),i(t))
    =
    s(t)^{1/2}
    \cdot
    \frac{1}{2}
    \left[
        1
        +
        \tanh
        \left(
            \frac{i(t)N_{\mathrm{pop}} - 3 \times 10^5}{0.05 N_{\mathrm{pop}}}
        \right)
    \right].
    \label{eq:true_xi}
\end{equation}
This closure represents a smooth, saturating quarantine response.
When the infectious population count is low, the transition into quarantine is small.
As infections increase, the quarantine rate rises, representing increased testing, self-isolation, or public-health response.
The hyperbolic tangent produces a smooth transition near $3 \times 10^5$ infectious individuals.
The dependence on $s(t)^{1/2}$ reduces the quarantine response as the susceptible pool is depleted.

\begin{figure}[h!]
    \centering
    \includegraphics[width=0.7\textwidth]{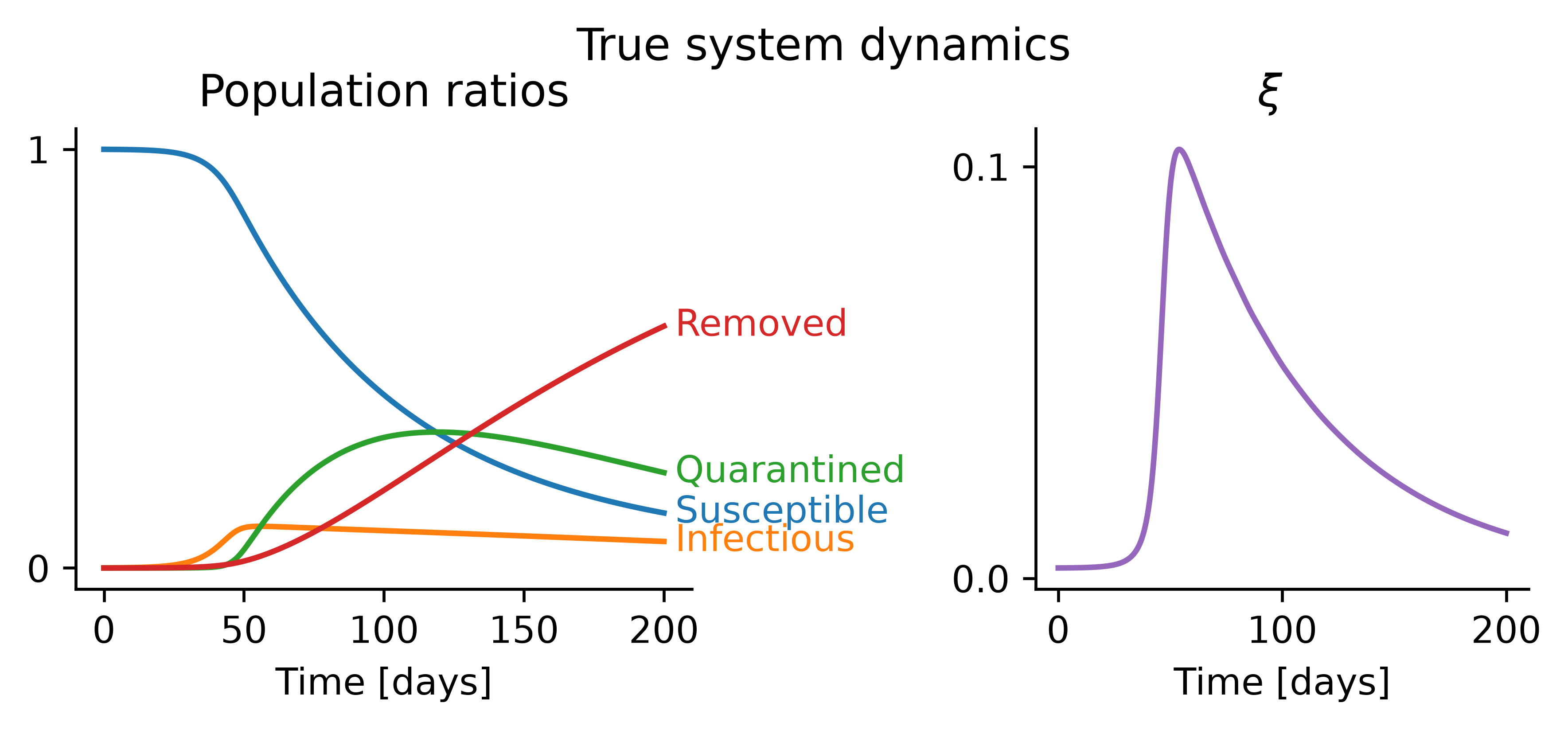}
    \vspace{-1em}
    \caption{True SIQR system dynamics and closure trajectory used to generate the synthetic observations.}
    \label{fig:true_system}
\end{figure}

For the dataset used in training, the population size is $N_{\mathrm{pop}}=2 \times 10^6$.
The initial counts are $500$ infectious individuals, $10$ quarantined individuals, $10$ removed individuals, and the remainder susceptible.
The nondimensional initial condition is therefore
\begin{align}
    i(0) &= \frac{500}{N_{\mathrm{pop}}},
    \label{eq:initial_condition_i}
    \\
    q(0) &= \frac{10}{N_{\mathrm{pop}}},
    \label{eq:initial_condition_q}
    \\
    r(0) &= \frac{10}{N_{\mathrm{pop}}},
    \label{eq:initial_condition_r}
    \\
    s(0) &= 1-i(0)-q(0)-r(0).
    \label{eq:initial_condition_s}
\end{align}
The transition rate parameters in the model are set at $\beta=0.15$, $\gamma=0.013$, and $\delta=0.01$.
The time evolution of this system is shown in~\Cref{fig:true_system}.

Noisy observations are generated as
\begin{equation}
    d_i
    =
    H x(t_i)
    +
    \epsilon_i,
    \qquad
    \epsilon_i \sim \mathcal{N}(0,\sigma_{\epsilon}^2 I),
    \label{eq:siqr_noisy_obs}
\end{equation}
with $H=I\in\mathbb{R}^4$.
The measurement noise is assumed known and is set to $\sigma_{\epsilon}^2=10^{-6}$.

\subsubsection{SIQR embedded closure model}
\label{sec:siqr_embedded}

The embedded approach replaces the unknown quarantine closure with a neural network $\xi_{\theta}$.
The embedded SIQR dynamics are
\begin{align}
    \frac{d s}{d t}
    &=
    -\beta s i,
    \label{eq:embedded_siqr_s}
    \\
    \frac{d i}{d t}
    &=
    \beta s i - \gamma i - \xi_{\theta}(s,i,q,r)i,
    \label{eq:embedded_siqr_i}
    \\
    \frac{d q}{d t}
    &=
    \xi_{\theta}(s,i,q,r)i - \delta q,
    \label{eq:embedded_siqr_q}
    \\
    \frac{d r}{d t}
    &=
    \gamma i + \delta q.
    \label{eq:embedded_siqr_r}
\end{align}
The positive sign in the quarantined equation reflects that the learned closure transfers individuals from the infectious compartment into the quarantined compartment.
The state variable trajectory induced by a given neural network parameter vector \(\theta\) is denoted by \(x(t;\theta)\).
The network is trained against noisy state observations using the embedded trajectory loss described in Section~\ref{sec:methodology_training_objectives}.
Training, validation, random-start selection, and local-refinement accept-or-reject decisions follow the comparison protocol in Section~\ref{sec:methodology}.
The specific optimizer, initialization, and solver settings used in this study are given in Section~\ref{sec:numerical_implementation}.

\subsubsection{SIQR staged data-assimilated closure model}
\label{sec:siqr_staged}

For the staged data-assimilated approach, the SIQR state is augmented with a time-dependent latent closure variable.
The augmented state is
\begin{equation}
    \hat{z}(t)
    =
    \begin{bmatrix}
        s(t) & i(t) & q(t) & r(t) & \hat{\xi}(t)
    \end{bmatrix}^T.
    \label{eq:siqr_augmented_state}
\end{equation}
The augmented dynamics used during data assimilation are
\begin{align}
    \frac{d s}{d t}
    &=
    -\beta s i,
    \label{eq:staged_siqr_augmented_s}
    \\
    \frac{d i}{d t}
    &=
    \beta s i - \gamma i - \hat{\xi}(t)i,
    \label{eq:staged_siqr_augmented_i}
    \\
    \frac{d q}{d t}
    &=
    \hat{\xi}(t)i - \delta q,
    \label{eq:staged_siqr_augmented_q}
    \\
    \frac{d r}{d t}
    &=
    \gamma i + \delta q,
    \label{eq:staged_siqr_augmented_r}
    \\
    d\hat{\xi}(t)
    &=
    \sigma_{\xi} dW(t).
    \label{eq:staged_siqr_augmented_xi}
\end{align}
The EnKF ensemble size is $N_{\mathrm{ens}}=100$.
The closure process noise is applied only to the closure component.
The observation covariance is $\Gamma=\sigma_{\epsilon}^2 I_4$.
The assimilation uses observations indexed by $\mathcal{I}_{\mathrm{assim}}=\mathcal{I}_{\mathrm{train}}\cup\mathcal{I}_{\mathrm{val}}$.
The original test observations indexed by $\mathcal{I}_{\mathrm{test}}$ are not assimilated.

The EnKF and fixed-interval smoother produce smoothed ensemble means $\bar{z}_i^s = [\bar{x}_i^s,\bar{\xi}_i^s]^T$.
For the SIQR problem, the smoothed physical-state mean is $\bar{x}_i^s = [\bar{s}_i^s,\bar{i}_i^s,\bar{q}_i^s,\bar{r}_i^s]^\top$.
The staged neural network is trained offline using the supervised loss
\begin{equation}
    \mathcal{L}_{\mathrm{staged}}(\theta)
    =
    \frac{1}{2}
    \sum_{i \in \mathcal{I}_{\mathrm{train}}}
    \left|
        \xi_{\theta}(\bar{s}_i^s,\bar{i}_i^s,\bar{q}_i^s,\bar{r}_i^s)
        -
        \bar{\xi}_i^s
    \right|^2.
    \label{eq:staged_loss_siqr}
\end{equation}
The pseudo-validation loss is computed using the same expression over $\mathcal{I}_{\mathrm{val}}$.
After training, the learned staged closure is embedded into the SIQR equations for further predictions:
\begin{align}
    \frac{d s}{d t}
    &=
    -\beta s i,
    \label{eq:staged_rollout_siqr_s}
    \\
    \frac{d i}{d t}
    &=
    \beta s i - \gamma i - \xi_{\theta}(s,i,q,r)i,
    \label{eq:staged_rollout_siqr_i}
    \\
    \frac{d q}{d t}
    &=
    \xi_{\theta}(s,i,q,r)i - \delta q,
    \label{eq:staged_rollout_siqr_q}
    \\
    \frac{d r}{d t}
    &=
    \gamma i + \delta q.
    \label{eq:staged_rollout_siqr_r}
\end{align}

\subsection{Numerical implementation}
\label{sec:numerical_implementation}

The code used to generate the numerical results in this section is available at \url{https://github.com/sandialabs/STAGED_DDCM-Staged_Embedded_Comparison}.
All numerical experiments are implemented in Python using NumPy, SciPy, and packages in the JAX ecosystem.
Synthetic SIQR trajectories and noisy observation data are generated using SciPy's \texttt{solve\_ivp} method with the RK45 timestepper.
The true system is solved on the time interval \([0,200]\) days.
Observations are stored every \(\Delta t = 0.25\) days, giving \(801\) observation times including both endpoints.

The embedded and staged closure networks use the same architecture, which was selected through a manual search using the supervised training data from the staged approach.
Each network takes the four instantaneous state variables \((s,i,q,r)\) as input and has two hidden layers with two nodes per layer, sigmoid hidden-layer activations, and one scalar output.
The output layer is unconstrained.
Weights are initialized using Xavier uniform initialization and biases are initialized to zero.
Both approaches use \(10\) random starts.

Neural-network training uses a two-stage derivative-based local optimization procedure.
Each random start is first trained with AdamW, an adaptive first-order method~\cite{loshchilov_decoupled_2019}.
The resulting candidate is then refined using L-BFGS~\cite{liu_limited_1989} as a quasi-Newton local polishing step.
The L-BFGS-refined candidate is accepted only if it improves the relevant validation objective.
For embedded training, this validation objective is the raw-observation validation loss.
For staged training, this validation objective is the EnKF-derived pseudo-validation loss.

AdamW uses learning rate \(10^{-3}\), weight decay \(10^{-6}\), and gradient clipping with maximum global norm \(1\).
Each AdamW run is allowed at most \(2000\) iterations.
Validation is performed every \(100\) AdamW iterations with early-stopping patience of \(4\) validation checks, so training stops after at most \(400\) AdamW iterations without validation improvement.
After AdamW, L-BFGS is run for at most \(500\) iterations.

Embedded closure simulations are computed using Diffrax with the Dopri5 solver and an adaptive step size controller.
The same Diffrax solver settings are used when the trained staged closure is embedded into the SIQR equations for predictions after training.
A neural-network start is counted as failed if the ODE solve fails, produces nonfinite states or losses, or returns the large solver-failure penalty used by the training code.

The EnKF and smoother are implemented in custom NumPy code.
The EnKF ensemble size is \(N_{\mathrm{ens}}=100\).
The augmented SIQR forecast model uses a forward-Euler update with internal step size \(\Delta t_{\mathrm{int}}=0.25\) days.
The initial closure value in the augmented state is \(\hat{\xi}(0)=0\).
The closure process noise is applied only to the closure component of the augmented state.

For the staged approach, the EnKF assimilates only the original training and validation observations.
The original test observations are never assimilated.
The closure process-noise scale \(\sigma_\xi\) is estimated separately for each replicate using an EnKF likelihood computed from the training and validation observations.
For this likelihood optimization, every tenth training-or-validation observation is used.
The process-noise optimization uses SciPy's \texttt{minimize} with the Nelder--Mead method, initial value \(2.0\times 10^{-5}\), bounds \([10^{-8},0.5]\), and tolerance \(10^{-5}\).
After estimating \(\sigma_\xi\), a final EnKF assimilation is performed using all training and validation observations.
Smoothing is employed in the staged experiments.
The staged neural network is trained using only the EnKF-derived state--closure pairs and does not call the SIQR simulator during supervised optimization.

For each replicate in the study, independent random-number streams are used for observation noise, train--validation--test splitting, EnKF sampling, embedded neural-network initialization, and staged neural-network initialization.
Specifically, replicate-dependent seeds are used for observation noise, data splitting, EnKF sampling, embedded training starts, and staged training starts.

\subsection{Calibration comparison}
\label{sec:calibration_results}

The study uses \(50\) independent replicates.
Both methods are evaluated on the same held-out test observations for each replicate.
The reported test loss is the likelihood-scaled observation-space loss defined in Section~\ref{sec:methodology_metrics}.
Because the study is synthetic, we also report closure RMSE on the test indices.
This closure RMSE is a diagnostic metric and would not be available in a real closure-modeling application.

\begin{figure}[h!]
    \centering
    \includegraphics[width=0.8\textwidth]{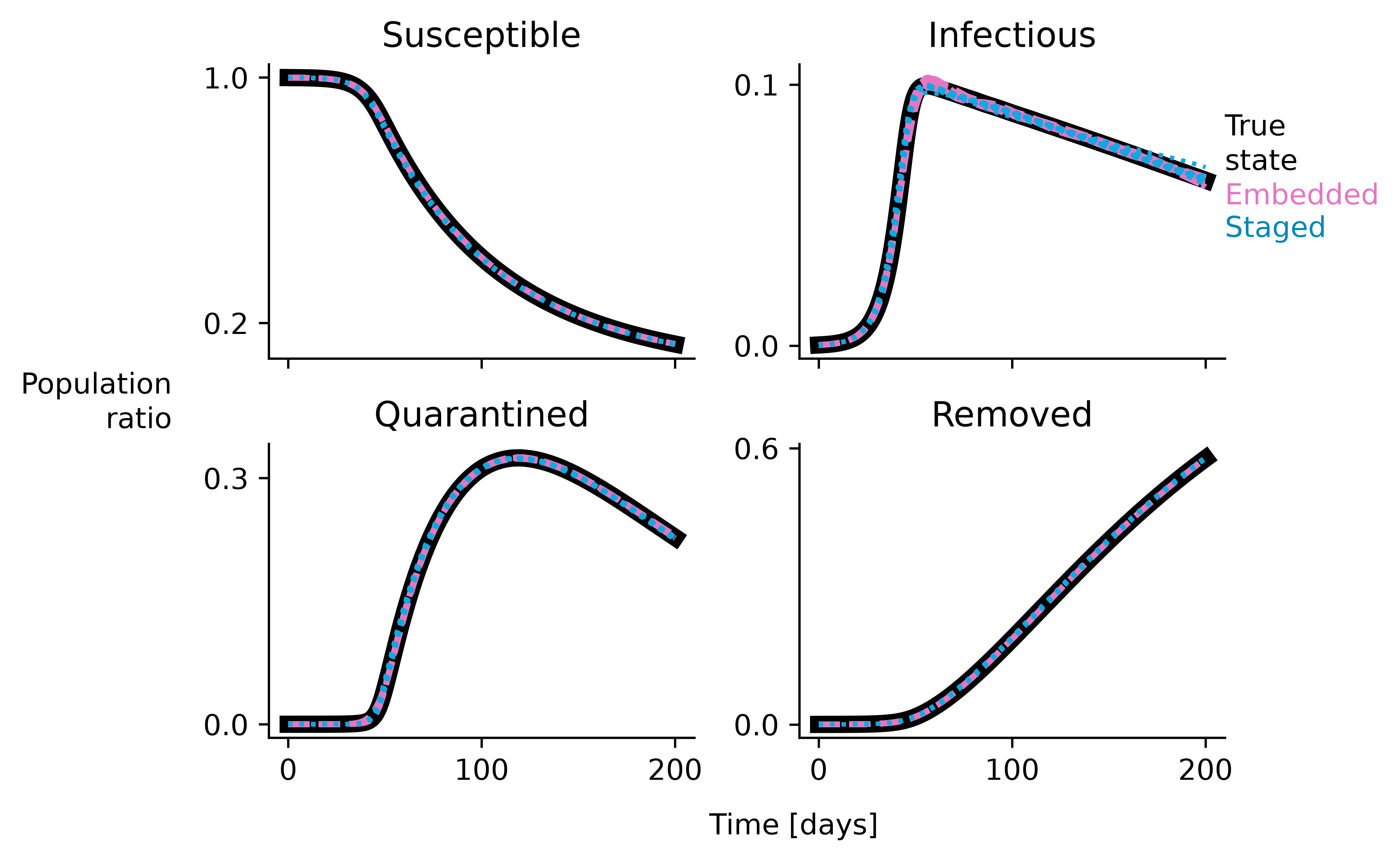}
    \vspace{-1em}
    \caption{Embedded and staged state trajectories across $50$ replicates compared with the true SIQR trajectory.}
    \label{fig:state_replicates}
\end{figure}

\begin{figure}[h!]
    \centering
    \includegraphics[width=0.5\textwidth]{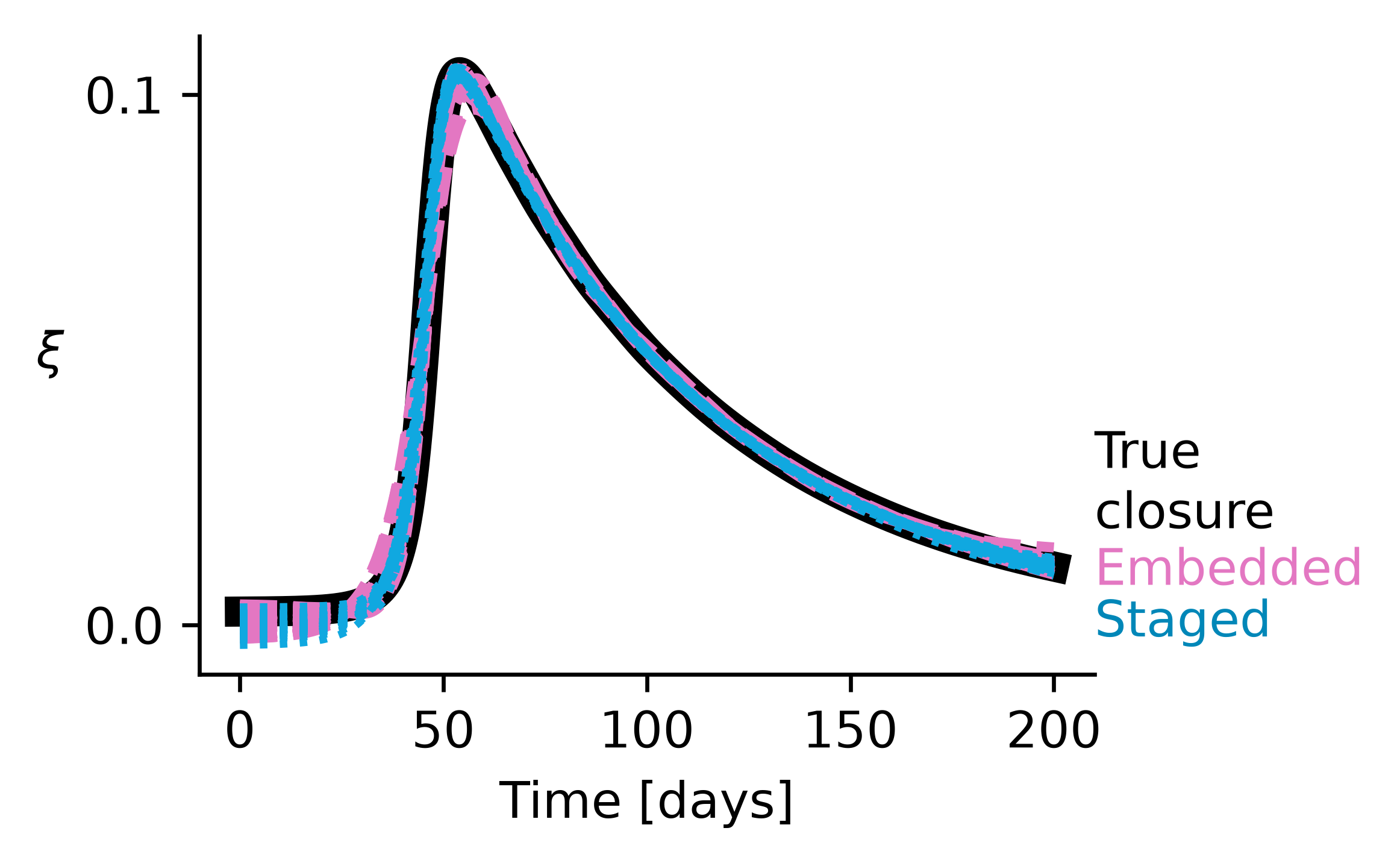}
    \vspace{-1em}
    \caption{Learned closure trajectories from embedded and staged replicates compared with the true synthetic closure.}
    \label{fig:closure_replicates}
\end{figure}

Figure~\ref{fig:state_replicates} compares the learned state trajectories from the embedded and staged approaches across replicates.
Both methods reproduce the qualitative SIQR dynamics quite well.
Figure~\ref{fig:closure_replicates} compares the learned closure trajectories with the true synthetic closure.
Again, both methods recover the broad temporal structure of the closure.

\begin{figure}[h!]
    \centering
    \includegraphics[width=0.5\textwidth]{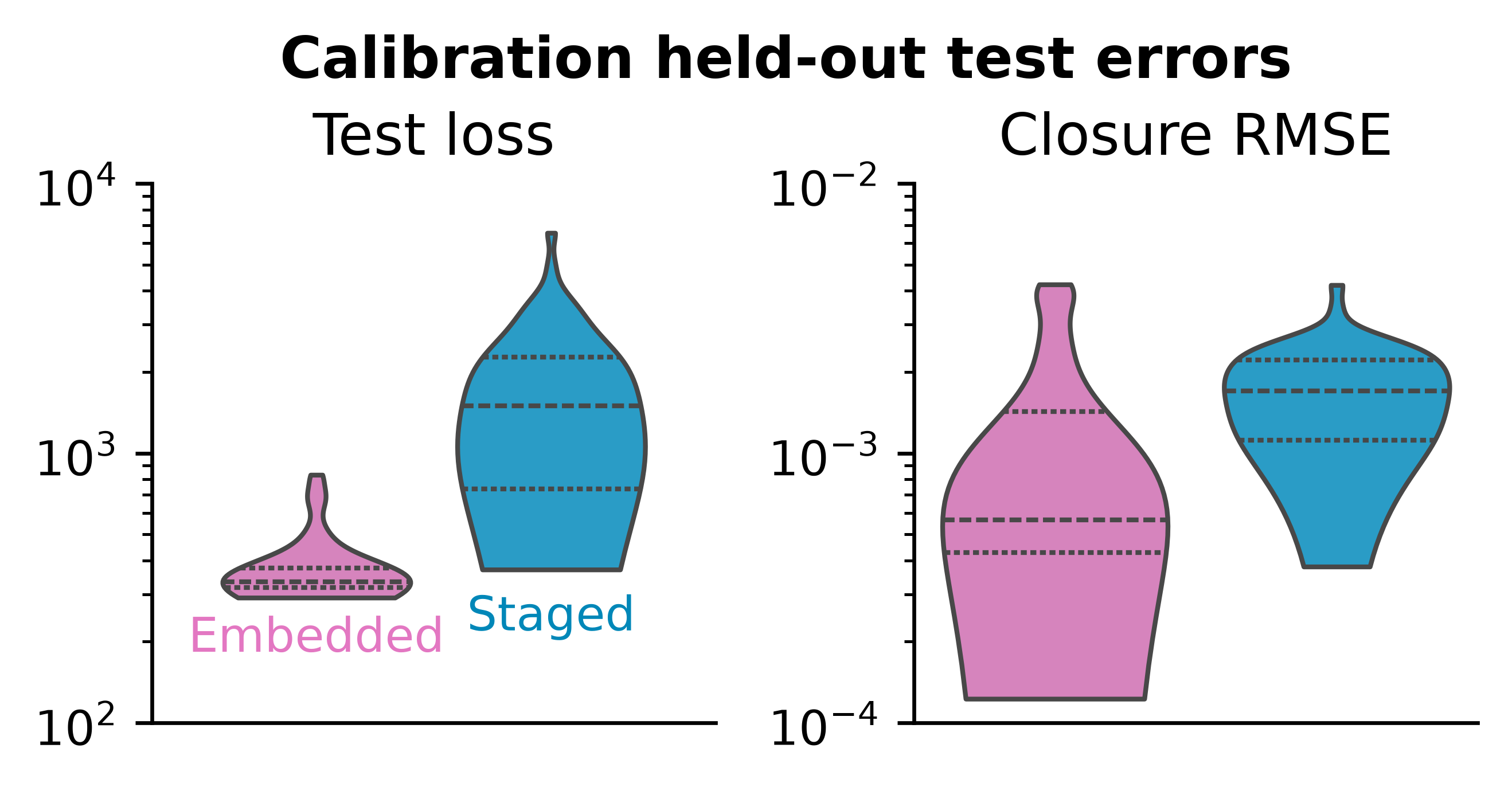}
    \vspace{-1em}
    \caption{Distribution of calibration errors over the $50$ independent replicates for the embedded and staged approaches.
    Each violin shows the empirical distribution of the replicate-level error metric for one method, with width indicating relative density.
    Horizontal lines inside each violin indicate the $25^{\rm th}$,  $50^{\rm th}$, and $75^{\rm th}$ percentiles of the replicate distribution.}
    \label{fig:test_errors}
\end{figure}

Figure~\ref{fig:test_errors} summarizes calibration accuracy across the 50 replicates, where calibration performance is reported using held-out observation-space test loss and synthetic closure RMSE on the calibration test indices.
The embedded approach achieves the lower average held-out test loss in this example, with mean test loss $3.92\times 10^2 \pm 1.35\times 10^2$.
The staged approach has a larger mean held-out test loss of $1.75\times 10^3 \pm 1.23\times 10^3$.
The embedded approach also has a lower average calibration closure RMSE of $1.18\times 10^{-3} \pm 1.17\times 10^{-3}$, compared with $1.70\times 10^{-3} \pm 7.42\times 10^{-4}$ for the staged approach.
Although the embedded approach is more accurate on average, the staged errors remain within the same order of magnitude for this calibration problem.
Thus, the calibration results indicate a measurable but not prohibitive loss in average accuracy for the staged approach.

The computational-cost and robustness comparison presented in Table~\ref{tab:calibration_summary} favors the staged approach.
The embedded approach requires $2.41\times 10^4 \pm 7.41\times 10^3$ training-stage model evaluations.
The staged approach requires $2.78\times 10^3 \pm 3.59\times 10^2$ training-stage model evaluations.
This corresponds to an approximately $9\times$ reduction in model evaluations.
The robustness comparison also favors the staged approach.
The embedded approach has $215$ failed neural-network starts across the $50$ replicates with $10$ multi-starts each, corresponding to a roughly 40\% failure rate, whereas the staged approach has no failed starts.

\begin{table}[h!]
    \centering
    \renewcommand{\arraystretch}{1.2}
\begin{tabular}{lcc}
\hline
Method & Training model evaluations & Failed starts \\
\hline
Embedded & $2.41\times 10^{4} \pm 7.41\times 10^{3}$ & 215 \\
STAGED DDCM & $2.78\times 10^{3} \pm 3.59\times 10^{2}$ & 0 \\
\hline
\end{tabular}
    \caption{Comparison of embedded and staged closure-training approaches over \(50\) replicates.
    Computational cost is measured by training-stage model evaluations, where results are presented as \(\mu \pm \sigma\) over the \(50\) replicates.
    Robustness is measured by the total number of failed neural-network starts.}
    \label{tab:calibration_summary}
\end{table}

As shown in Table~\ref{tab:staged_cost_breakdown}, the staged cost is dominated by process-noise maximum-likelihood estimation.
The process-noise MLE requires $2.68\times 10^3 \pm 3.59\times 10^2$ model evaluations, while the final EnKF assimilation requires 100 model evaluations.
The supervised staged neural-network training is offline and does not invoke the SIQR simulator.
This result indicates that further efficiencies may be gained in the process-noise MLE step of the data assimilation process.

\begin{table}[h!]
    \centering
    \renewcommand{\arraystretch}{1.2}
\begin{tabular}{lc}
\hline
Component & Model evaluations \\
\hline
Process-noise MLE & \(2.68\times 10^{3}\) \( \pm \) \(3.59\times 10^{2}\) \\
Final EnKF assimilation & \(1.00\times 10^{2}\) \( \pm \) \(0\) \\
Total STAGED DDCM & \(2.78\times 10^{3}\) \( \pm \) \(3.59\times 10^{2}\) \\
\hline
\end{tabular}
    \caption{Training-stage model-evaluation accounting for the staged approach.
    The total is the sum of model evaluations used for process-noise maximum-likelihood estimation and the final EnKF assimilation.
    The subsequent supervised neural-network training step does not call the SIQR simulator and is therefore excluded from this simulator-dependent cost.}
    \label{tab:staged_cost_breakdown}
\end{table}

\subsection{Validation-scenario generalization comparison}
\label{sec:validation_generalization}

\begin{figure}[h!]
    \centering
    \includegraphics[width=0.4\textwidth]{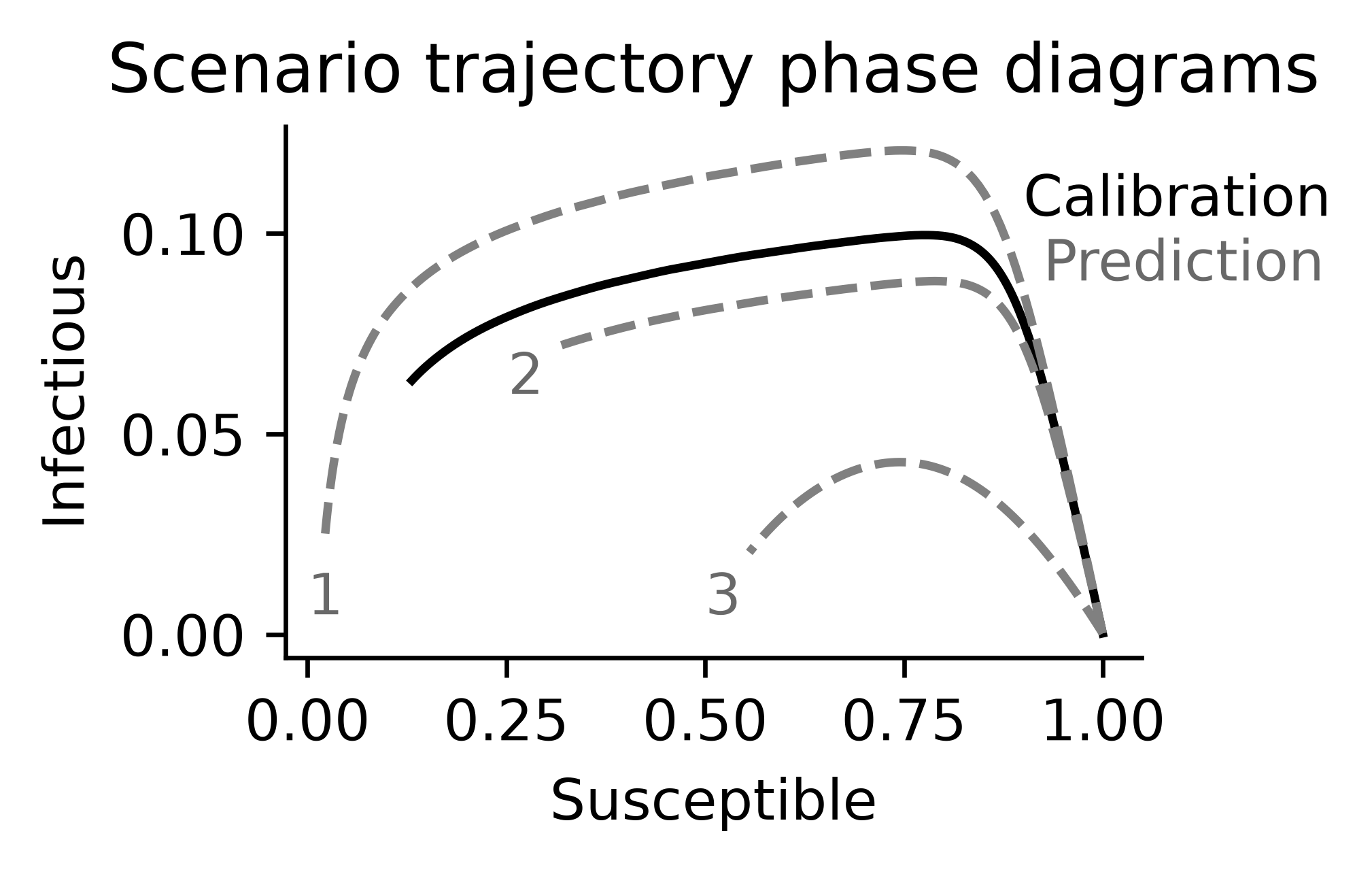}
    \vspace{-1em}
    \caption{Calibration and validation scenario trajectories in the susceptible-infectious projection.}
    \label{fig:phase_diagram}
\end{figure}

We also evaluate how the learned closures generalize when embedded into SIQR systems with parameter values different from the calibration scenario.
The validation scenarios are
\begin{align}
    \text{Case 1:} \quad
    & \beta=0.30,
    \quad
    \gamma=0.02,
    \quad
    \delta=0.01,
    \label{eq:validation_scenario_1}
    \\
    \text{Case 2:} \quad
    & \beta=0.10,
    \quad
    \gamma=0.01,
    \quad
    \delta=0.01,
    \label{eq:validation_scenario_2}
    \\
    \text{Case 3:} \quad
    & \beta=0.15,
    \quad
    \gamma=0.10,
    \quad
    \delta=0.01.
    \label{eq:validation_scenario_3}
\end{align}
These cases and their relation to the calibration scenario are illustrated through a susceptible-infectious phase diagram in Figure~\ref{fig:phase_diagram}.
Case 1 is a more transmissible scenario with a shorter infectious period.
Case 2 remains closer to the calibration trajectory and is closer to an interpolation regime in state space.
Case 3 induces stronger extrapolation through a substantially accelerated recovery rate.

\begin{figure}[h!]
    \centering
    \includegraphics[width=0.8\textwidth]{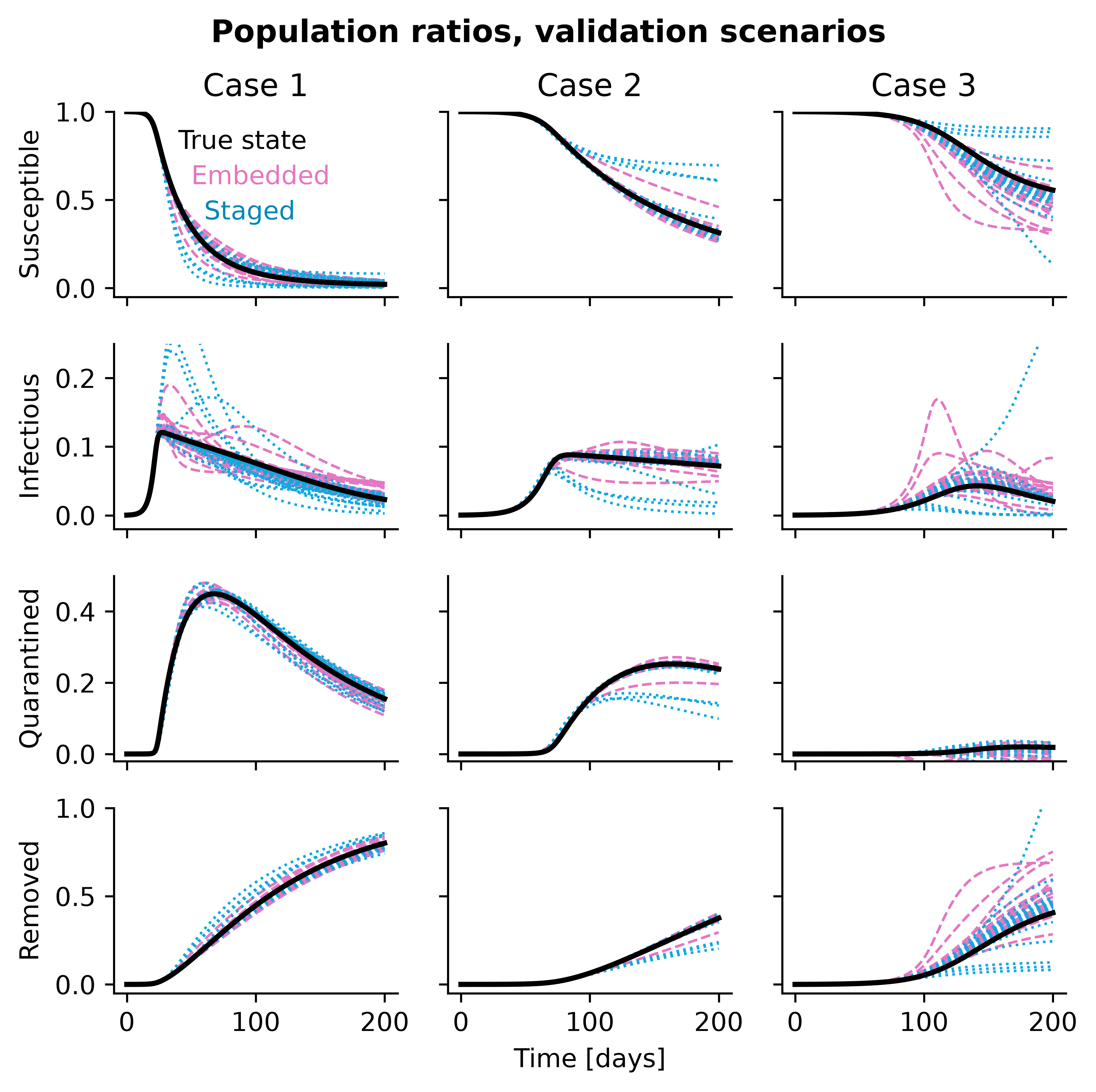}
    \vspace{-1em}
    \caption{Embedded and staged state trajectories under validation scenarios.}
    \label{fig:validation_states}
\end{figure}

\begin{figure}[h!]
    \centering
    \includegraphics[width=0.7\textwidth]{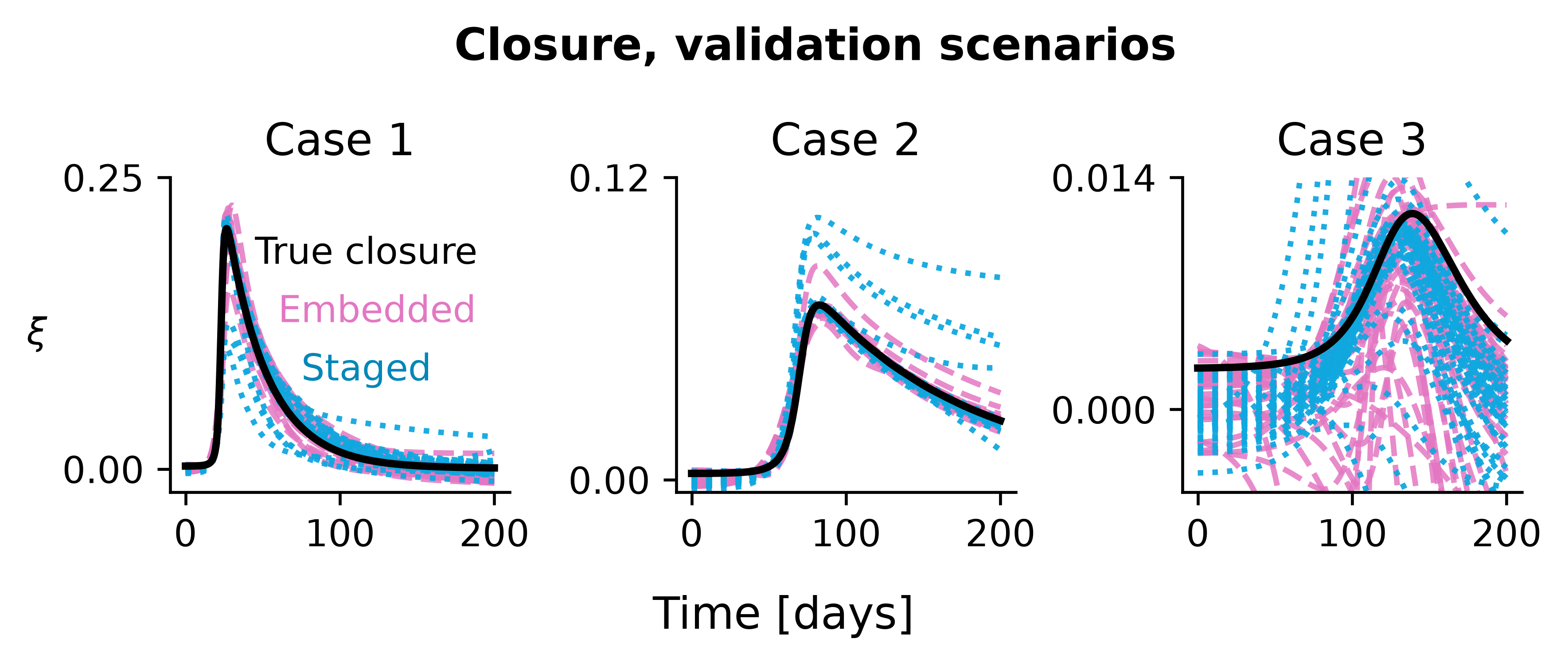}
    \vspace{-1em}
    \caption{Embedded and staged learned closure trajectories under validation scenarios.}
    \label{fig:validation_closures}
\end{figure}

Figures~\ref{fig:validation_states} and~\ref{fig:validation_closures} show state and closure trajectories for the validation scenarios.
Both embedded and staged closures perform similarly across scenarios, with the greatest qualitative variability in results for Case 3, which constitutes the greatest extrapolation from the calibration scenario.
Qualitative variability seems similar for the other two cases, which are closer to the calibration scenario.

\begin{figure}[h!]
    \centering
    \includegraphics[width=0.8\textwidth]{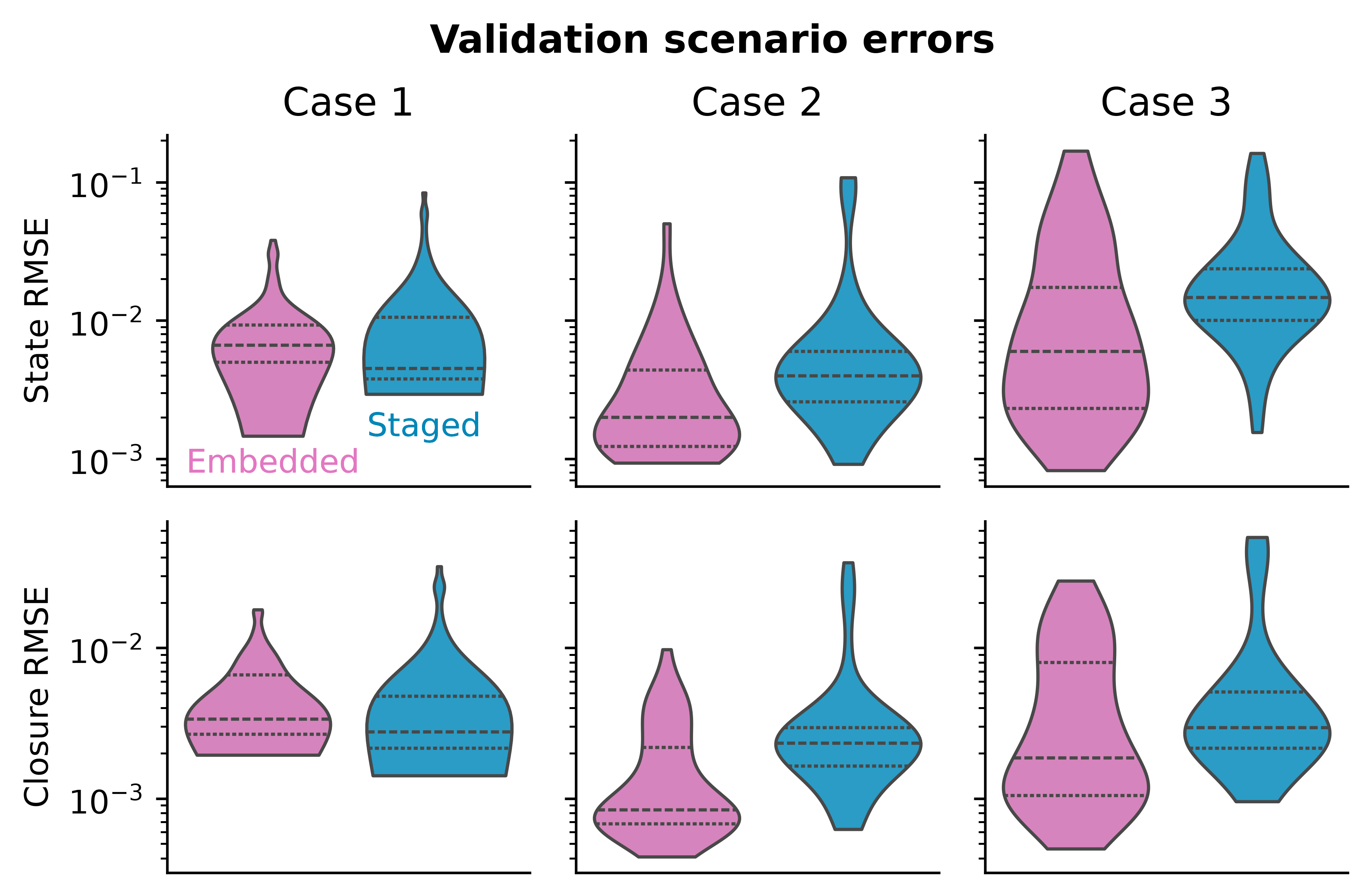}
    \vspace{-1em}
    \caption{Distribution of validation scenario errors over the $50$ independent replicates for the embedded and staged approaches.
    Each violin shows the empirical distribution of the replicate-level error metric for one method, with width indicating relative density.
    Horizontal lines inside each violin indicate the $25^{\rm th}$,  $50^{\rm th}$, and $75^{\rm th}$ percentiles of the replicate distribution.}
    \label{fig:validation_rmse}
\end{figure}

Figure~\ref{fig:validation_rmse} summarizes validation scenario accuracy across the 50 replicates.
The validation metrics are consistent with the calibration results but also show that the staged approach can remain competitive.
For Case 1, the embedded approach has mean state RMSE $9.43\times 10^{-3}$ and mean closure RMSE $5.09\times 10^{-3}$.
For the same case, the staged approach has mean state RMSE $1.15\times 10^{-2}$ and mean closure RMSE $5.74\times 10^{-3}$.
These values are slightly larger for the staged approach but are close in magnitude, and the median errors in Case 1 are favorable to the staged approach.
For Case 2, the embedded approach has lower mean and median state and closure errors.
For Case 3, both methods become less accurate, and the embedded approach again has lower average errors.
Overall, the validation scenarios suggest that staged training can produce useful closures outside the calibration trajectory, but with greater replicate-to-replicate variability than embedded training in this example.

\subsection{Discussion}

Taken together, these results show an accuracy--efficiency tradeoff in the SIQR exemplar.
The embedded approach is more accurate on average for this inexpensive low-dimensional system, but the staged approach attains the same qualitative behavior with substantially fewer model evaluations and no failed neural-network starts.
The staged approach therefore sacrifices some average accuracy in exchange for lower computational cost and improved optimization robustness.

Whether this sacrifice is worthwhile depends on the computational setting.
For inexpensive differentiable models, embedded training may be preferable.
For expensive or nondifferentiable models, or for models where numerical instability can be easily induced, the reduced model-evaluation cost and improved robustness of staged training may be more important than the moderate increase in error observed here.
Because this comparison is based on one synthetic SIQR example, the numerical values should not be interpreted as universal.
Instead, they illustrate the type of accuracy--cost trade-off that can arise when replacing embedded neural-closure training with a staged data-assimilated workflow.

\section{Conclusions}
\label{sec:conclusions}

This work compared embedded and staged data-assimilated approaches for learning neural closure models in dynamical systems, focusing on accuracy, computational cost, and optimization robustness.
Embedded training optimizes the closure through trajectory-level mismatch, whereas staged training uses data assimilation to infer closure-consistent trajectories before fitting the closure through offline supervised regression.

In the synthetic SIQR example considered here, embedded training produced the lowest average held-out calibration error and the lowest average diagnostic closure error.
However, the staged approach produced qualitatively similar trajectories and closure behavior with errors that were often comparable in magnitude.
Thus, the results should be interpreted as an accuracy--efficiency tradeoff rather than as a definitive ranking of the two approaches.
For this low-dimensional and inexpensive exemplar, the accuracy advantage of embedded training may justify its additional computational cost.
For more expensive, nondifferentiable, legacy, or numerically fragile simulators, the reduced model-evaluation cost and improved robustness of staged training may be more important than the moderate increase in error observed in this example.

The computational and robustness differences were substantial.
Across $50$ independent replicates, staged training required approximately $9\times$ fewer training-stage model evaluations and eliminated failed neural-network starts.
In contrast, embedded training yielded an approximately 40\% training failure rate due numerical challenges caused by intermediate closure model parameter guesses leading to instability, solver failures, etc.
Staged training avoids this issue by separating computational model evaluations during data assimilation from offline neural-network optimization.

Both approaches produced generally stable predictions under parameter perturbations.
Embedded training generally achieved lower mean state and closure errors, while staged training remained competitive in some scenarios and captured meaningful features of the nonlinear quarantine response.
The larger staged mean errors were partly driven by outliers, indicating greater replicate-to-replicate variability.
These findings suggest that staged learning can generalize beyond the calibration trajectory, although prediction errors may accumulate when trajectories depart from the calibration manifold.

Future work should extend the comparison to more complex and high-dimensional systems, incorporate uncertainty in the inferred closure, and use multiple calibration scenarios to improve generalization.
Further investigation should also account for architecture selection and hyperparameter tuning, particularly the potential to reuse assimilated datasets for offline model development without additional simulator evaluations.
These extensions will help establish when the computational and robustness advantages of staged training outweigh the accuracy gains of embedded optimization.

\section*{Acknowledgments}
Sandia National Laboratories is a multi-mission laboratory managed and operated by National Technology \& Engineering Solutions of Sandia, LLC (NTESS), a wholly owned subsidiary of Honeywell International Inc., for the U.S. Department of Energy's National Nuclear Security Administration (DOE/NNSA) under contract DE-NA0003525. This written work is authored by an employee of NTESS. The employee, not NTESS, owns the right, title and interest in and to the written work and is responsible for its contents. Any subjective views or opinions that might be expressed in the written work do not necessarily represent the views of the U.S. Government. The publisher acknowledges that the U.S. Government retains a non-exclusive, paid-up, irrevocable, world-wide license to publish or reproduce the published form of this written work or allow others to do so, for U.S. Government purposes. The DOE will provide public access to results of federally sponsored research in accordance with the DOE Public Access Plan.
This paper describes objective technical results and analysis. Any subjective views or opinions that might be expressed in the paper do not necessarily represent the views of the U.S. Department of Energy or the United States Government.
This work was supported by the Laboratory Directed Research and Development program (project
233077) at Sandia National Laboratories, a multimission laboratory managed and operated by National
Technology and Engineering Solutions of Sandia LLC, a wholly owned subsidiary of Honeywell International
Inc. for the U.S. Department of Energy’s National Nuclear Security Administration under contract DE-
NA0003525.

\bibliographystyle{unsrt}
\bibliography{references}
\end{document}